\documentclass[journal]{IEEEtran}
\usepackage[T1]{fontenc}
\usepackage{cite}
\usepackage{url}
\usepackage{algorithmic}
\usepackage{array}
\usepackage{amsmath,amssymb,amsfonts}
\usepackage{standalone}
\usepackage{tikz}
\usetikzlibrary{positioning}     
\usepackage{hyperref}
\usepackage{booktabs}
\usepackage{overpic}
\usepackage{rotating}
\usepackage{enumitem}
\usepackage{balance}
\usepackage{algorithm}
\usepackage[caption=false,font=normalsize,labelfont=sf,textfont=sf]{subfig}
\usepackage{textcomp}
\usepackage{stfloats}
\usepackage{verbatim}
\usepackage{graphicx}

\newcommand{\TT}[1]{\protect\scalebox{0.75}[1.04]{\texttt{#1}}}

\begin{document}
	\title{How to Do Machine Learning with Small Data? \\-- A Review from an Industrial Perspective}
	
	\author{Ivan Kraljevski, Yong Chul Ju, Dmitrij Ivanov, Constanze Tsch{\"o}pe, and Matthias Wolff
		\thanks{This study was supported by the Ministry of Science, Research and Culture of Brandenburg through the Project Kognitive Materialdiagnostik under Grant 22-F241-03-FhG/007/001. \textit{\{Corresponding author: Ivan Kraljevski.\}}
		}
		\thanks{Ivan Kraljevski, Yong Chul Ju and Constanze Tsch{\"o}pe are with Fraunhofer Institute for Ceramic Technologies and Systems, Fraunhofer IKTS, Dresden, Germany (e-mail: ivan.kraljevski@ikts.fraunhofer.de).}
		\thanks{Dmitrij Ivanov, Matthias Wolff are with Brandenburg University of Technology Cottbus--Senftenberg, Cottbus, Germany (e-mail: matthias.wolff@b-tu.de).}
	}
	
	\maketitle
	
	\begin{abstract}
		Artificial intelligence experienced a technological breakthrough in science, industry, and everyday life in the recent few decades. The advancements can be credited to the ever-increasing availability and miniaturization of computational resources that resulted in exponential data growth.
		However, because of the insufficient amount of data in some cases, employing machine learning in solving complex tasks is not straightforward or even possible.
		As a result, machine learning with small data experiences rising importance in data science and application in several fields. 
		The authors focus on interpreting the general term of ``small data'' and their engineering and industrial application role. They give a brief overview of the most important industrial applications of machine learning and small data. Small data is defined in terms of various characteristics compared to big data, and a machine learning formalism was introduced. 
		Five critical challenges of machine learning with small data in industrial applications are presented: unlabeled data, imbalanced data, missing data, insufficient data, and rare events. 
		Based on those definitions, an overview of the considerations in domain representation and data acquisition is given along with a taxonomy of machine learning approaches in the context of small data.
	\end{abstract}
	
	

	
	\section{Introduction}\label{sec:Intro}
	
	Machine learning became the most popular \mbox{buzzword} nowadays in business meetings and is often propagated as the magic novel cure for all problems and the carrier of the digital revolution. However, the general term machine learning contains a variety of data-driven methods which are fundamental for ``artificial intelligence'' (AI) \cite{BHRCADHT17}, coined by John McCarthy \cite{RJ16} and pushed by Alan Turing's question ``can machine think?'' to the modern AI research known today \cite{T50, JM15}. In different contexts, machine learning (ML) is also known as data mining, or predictive analysis \cite{D12}. 
	
	T.\,M. Mitchel \cite{MT97} defines machine learning as: ``A computer program is said to \emph{learn} from experience $E$ concerning some class of tasks $T$ and performance measure $P$, if its performance at tasks in $T$, as measured by $P$, improves with experience $E$''. Here, classes of tasks $T$ encompass different types of prediction and inference, the source of experience $E$ is the available ``a priori'' knowledge about the domain, including domain-specific data samples, and the $P$ is a measure of success achieved in a particular task $T$.
	A less formal definition is given in \cite{C16}, \emph{machine learning} (ML) ``is a sub-field of computer science, but is often also referred to as predictive analytics, or predictive modeling''. Its goal and usage is to build new and/or leverage existing \emph{algorithms} to \emph{learn} from data, in order to \emph{build generalizable models} that give \emph{accurate predictions}, or to find \emph{patterns}, particularly with \emph{new and unseen similar data}''.
	
	In the last two decades, considerable progress was made in ML-science leveraging the growing computational power \cite{JM15, LBH15, S15} and the vast amounts of data increasingly being available (big data), see \cite{RGR18}. 
	Consequently, data-driven ML methods are applied to social, economic, industrial, environmental, and even political tasks \cite{HWJ20}. A large variety of comprehensive ML publications appeared, e.g.,\cite{D12, JM15, BHRCADHT17} with references therein related to one such as search engines, machine translation, online retailers, social networking services, financial modeling, marketing, education, and policy-making. Further detailed understanding of the underlying mechanisms of ML methods was gained as in statistical learning theory \cite{Va95, HTF09}, pattern classification \cite{DHS00, B06}, neural and deep learning (DL) \cite{KIH12, LBH15, S15, GBC16} in different applications.
	
	\paragraph*{Small data} The adoption of big data-driven ML methods is \emph{not} always feasible, especially when the amount of available data is insufficient for the desired performance due to various limitations or when data-sets are high-dimensional, complicated, or expensive. 
	For example, cases such as medical magnetic resonance therapy scans of new or growing cancer cells, expensive and time-consuming aero-engine test data, limited data, or even a lack of knowledge about unknown datasets, shall be covered by the term ``small data'' as a counterpart to ``big data''. 
	The term ``small data'' was promoted by many authors and had different interpretations, e.g. ``Small data results from the experimental or intentionally collected data of a human scale where the focus is on causation and understanding rather than prediction'' \cite{FA18}. In \cite{LM29016} the author remarks ``a lone piece of small data is almost never meaningful enough to build a case or create a hypothesis, but blended with other insights and observations ... comes together to create a solution that forms the foundation of a future brand or business''.
	Other interpretations of small data is presented in review publications, e.g.\cite{LDWLW18,QL20}, neural network small data approaches \cite{OWB18} and sensor fusion applications \cite{VBMS18}.
	
	While small data are already widespread in social and financial applications, they are relatively rare in industrial and engineering. The availability, quality, and composition of the data significantly influence the performance. ML should be able to address the main challenges, such as, to handle high-dimensional problems and data-sets, to present transparent and concrete results, to adapt to dynamic environmental conditions, to expand the knowledge by learning from results, to utilize available data without special treatment, to identify inter-and intra-relations in the processes, correlation, casualty \cite{WWIT16, WYLR18}.
	
	\subsection{Industrial applications} They encompass all the processes, directly and indirectly, related to the manufacturing, or maintenance of manufacturing plants, such as, in automotive, aviation, chemical, pharmaceutical, food, and steel industry. Machine learning methods are already widely applied in various industries for process optimization, monitoring, and control applications \cite{GSDSB17, LADOPB18, MAYR2019, MVPCDAF22}. In semiconductor manufacturing \cite{LKK17}, rotational machinery \cite{LYZC18}, materials degradation \cite{NDB18}, engine failure prediction \cite{NNDBB20} as some examples in predictive maintenance.
	
	\subsection{Engineering applications} These applications cover all the fields which apply scientific principles for developing, designing, building, and analyzing technological solutions, such as bio-engineering, e.g., modeling and identification of gene and protein structures \cite{YWA19}, DNA research \cite{ZQJ19}, predictive modeling in biomedical engineering \cite{SLDBHK15}, drug discovery \cite{ARPP17,VCCDFLMSS19}; in electrical engineering, e.g., autonomous driving \cite{BTDFFGJMMZZZ16}; in civil engineering, e.g., structural design optimization, structural health monitoring \cite{SB18, HCZG21}; in mechanical engineering, e.g., flow field prediction \cite{SMPW19}, data-driven flow dynamics \cite{TWPH18}, data-driven computational mechanics \cite{KO16}; in material science\cite{ZL18, SMBM19, PG21}, e.g., prediction of material properties \cite{CC2020}, materials design and discovery \cite{GL18, BACHKK19}; in optimization tasks for complex and expensive black-box problems, e.g., smart data \cite{GR10, I15}.
	
	Despite the rapid growth and availability of big data, small data play an even more critical role in applying ML in industrial and engineering. Small data allows a more focused ML design providing answers to more targeted questions and explainable models. ML with big data is applicable only if the domain of analysis is expected to be stationary, whereas the dynamics of systems often change over time.
	
	Also, in healthcare and biology sciences, the data is obtained via experimentation, and the formulation of relevant hypotheses to explain the data is the limiting factor for employing approaches based on big data.
	Therefore we want to divide the general term ``small data'' into different categories, which may partly overlap but may serve as guidelines to present the main ideas and ML techniques in the context of small data.
	
	It is important to emphasize that applying machine learning to small data for a specific domain also requires specific knowledge about the problem or the task.
	We could roughly divide the ML approaches into generalized and specialized according to the available data's nature. A straightforward application of ML frameworks without underlying knowledge may lead to wrong conclusions \cite{WR16}. That is why we do not claim to present the best-practice approaches, but inspire the interested reader to explore their solutions.
	
	\paragraph{Organization of the paper} The paper is organized as follows:
	In Section~\ref{sec:ChSD}, we first define \emph{small data} in terms of various characteristics compared with those of \emph{big data}, and we present a machine learning formalism which used throughout the paper. 
	
	Then, we describe five main challenges of small data ML as seen from industrial applications: unlabeled data, imbalanced data, missing data, insufficient data, and rare events. Based on those definitions, in Section~\ref{sec:MLwSD} we give the overview of the considerations in domain representation acquisition and taxonomy of machine learning approaches in the context of small data. 
	
	In Section~\ref{sec:AtSD}, we list some of the possible approaches to small data ML in terms of the five categories mentioned above. We close this paper with concluding remarks in Section~\ref{sec:Concl}.

	\section{Challenges of Small Data}\label{sec:ChSD}
	
	\subsection{Definition of Small Data}\label{ssec:DefSD} 
	To explore the general term ``small data'', we should first describe the characteristics of big data \cite{BC12, H16} in contrast with the small data \cite{KM16}:
	\begin{itemize}
		\item \emph{Volume} -- sheer quantity of the data, small data could have limited to large volume, where big data always have an enormous volume.	
		\item \emph{Variety} -- small data are are constrained in terms of their breadth and diversity, while big data draw insights from a wide spectrum range of data types. 	
		\item \emph{Velocity} -- small data are generated much slower and less continually compared to big data, which are produced much faster in real-time.	
		\item \emph{Veracity} -- in big data, the data quality and the value can vary greatly, and it can be messy, noisy, contain uncertainty and errors. In contrast, small data are usually produced in a controlled manner.	
		\item \emph{Exhaustivity} -- an entire system is captured (big data), rather than being sampled (small data).	
		\item \emph{Resolution} and \emph{indexicality} -- big data are fine-grained in resolution and uniquely indexical in identification, while small data resolution range from coarse to fine and in indexicality from weak to strong.	
		\item \emph{Relationality} -- big data contain common attributes that enable conjoining of different datasets, where the relationality in small data is weak.	
		\item \emph{Extensionality} and \emph{scalability} -- the big data can add or change new attributes easily and rapidly expand in size, where small data are difficult to administer and have limited extensionality and scalability.   
		\item \emph{Value} -- in big data, many insights can be extracted and the data repurposed, where small data have limited scope and rarely could be reused for different purposes.	
		\item \emph{Variability} -- the meaning of the data can constantly be shifting depending on the context in which they are generated, and this does not apply to small data which are quite inflexible regarding their generation.
	\end{itemize}
	
	In the case of industrial and engineering application, we must not draw a clear line to big data but motivate use cases or applications which exhibit some of the aforementioned attributes indicating small data. 
	Hence, we suggest scenarios that might include certain redundancies but serve as essential keywords: unlabeled, imbalanced, missing and insufficient data, and rare events.

	\subsection{Machine Learning Formalism and Data Sets}\label{ssec:MLFandDS}
	We start with introducing symbols and definitions, which we will use throughout the
	rest of this paper. The presented notation is loosely inspired by \cite{SS14, DFO20}.
	
	The main objective of machine learning is to find a map 
	\begin{equation}
	h:\mathcal{X}\to\mathcal{Y}
	\end{equation}
	from an \emph{input domain} $\mathcal X$ to an \emph{output domain} $\mathcal Y$. 
	This map is also called a predictor (classification, regression). It is to suit a
	particular task and to improve with ``experience'' (cf. Mitchel's definition of
	machine learning in the introduction). 
	The objects in the input domain are also known as inputs, data points, or samples, usually in the form of vectors of feature variables (or features). 
	The objects in the output domain are also known as outputs, ground truth, targets, or labels.
	
	Both, the input and the output domain, can be either numeric or symbolic. We
	call a domain \emph{numeric}, if it consists of $d$-dimensional real or complex
	vectors or vector sequences, i.e.,
	\begin{equation}
	\mathcal X,\mathcal Y \subseteq \left(\mathbb C^d\right)^+\!\!,
	\end{equation} 
	where ``$+$'' is the Kleene-Plus.\footnote{Non-empty sequence of arbitrary 
		length} We call a domain \emph{symbolic}, if it consists of symbolic entities. A
	symbolic entity can be an atomic symbol, any symbolic structure---strings,
	graphs, etc.---or a set of the aforementioned. Additionally, symbolic entities
	may contain (some) numeric values and also weights: weighted graphs, weighted
	sets, etc.
	
	If at least one of the domains, the input or the output domain, is numeric, 
	we speak of \emph{numeric machine learning}. If both domains are symbolic we
	speak of \emph{symbolic machine learning}.
	
	For industrial applications, the ``experience'' to learn from is typically
	supplied in form of data sets. Let $\varepsilon$ denote the empty or
	non-existing label. Then a data set can be formally defined as a multi-set:
	\begin{equation}\label{eq:multiset}
	\mathcal{D} := \big\{(x^{(1)},y^{(1)}),\ldots,(x^{(N)},y^{(N)})\big\}
	\subseteq \mathcal{X}\times\big(\mathcal{Y}\cup\{\varepsilon\}\big)
	\end{equation}
	consisting of $N:=|\mathcal{D}|$ features-label pairs $(x,y)\in\mathcal{D}$
	called \emph{samples}. Samples with $y=\varepsilon$ are called \emph{unlabeled},
	samples with $y\neq\varepsilon$ are called \emph{labeled}.
	
	In the following, we assume the label domain to be finite, $\mathcal{Y} :=
	\{c_1, \ldots, c_K\}$ with $K$ labels $c_1,\ldots,c_K$ which we call
	\emph{classes}. Then $\mathcal{Y}$ induces a $K+1$ partition of the data set:
	\begin{alignat}{2}
	\mathcal{D}_0 :=& \big\{(x,y)\in\mathcal{D}:y=\varepsilon\big\} && \text{ unlabeled subset,}\nonumber\\
	\mathcal{D}_1 := &\big\{(x,y)\in\mathcal{D}:y=c_1\big\} && \text{ labeled with class $c_1$,}\nonumber\\
	&\vdots \nonumber\\
	\mathcal{D}_K :=& \big\{(x,y)\in\mathcal{D}:y=c_K\big\} && \text{ labeled with class $c_K$.}\label{eq:partition}
	\end{alignat}
	Here the $\mathcal{D}_i$ are pairwise disjoint and fulfill
	\begin{equation}
	\mathcal{D}_0 \cup \mathcal{D}_1 \cup\ldots\cup \mathcal{D}_K=\mathcal{D}.
	\end{equation}
	Here, the $\mathcal{D}_i$ are pairwise disjoint and fulfill
	\begin{equation}
	\mathcal{D}_0 \cup \mathcal{D}_1 \cup\ldots\cup \mathcal{D}_K=\mathcal{D}.
	\end{equation}
	We denote by $N_i :=|\mathcal{D}_i|$ with $i\in\{0,1,\ldots,K\}$ the size of the subsets.
	
	We additionally define:
	\begin{alignat}{2}
	\mathcal{D}_{\text{unlabeled}} &:= \mathcal{D}_0 && \text{ unlabeled subset}\nonumber\\
	\mathcal{D}_{\text{labeled}} &:= \bigcup\nolimits_{i=1}^K \mathcal{D}_i && \text{ labeled subset}
	\end{alignat}
	and denote by
	\begin{equation}
	N_{\text{unlabeled}}=|\mathcal{D}_{\text{unlabeled}}|=N_0
	\end{equation}
	the number of unlabeled samples and by 
	\begin{equation}
	N_{\text{labeled}} =
	|\mathcal{D}_{\text{labeled}}| = \sum\nolimits_{i=1}^K N_i
	\end{equation}
	the number of labeled samples. We call a data set
	\begin{alignat}{2}
	\notag
	& \text{completely labeled, } &\quad & \text{if }
	N_{\text{unlabeled}}=0,
	\\
	\notag
	& \text{unlabeled, } && \text{if } N_{\text{labeled}}=0, 
	\text{ and}
	\\
	\notag
	& \text{partially labeled, } && \text{if } 0<N_{\text{labeled}}<N.
	\end{alignat}
	
	\begin{figure*}[!h]
		\centering
		\includegraphics[width=1\linewidth]{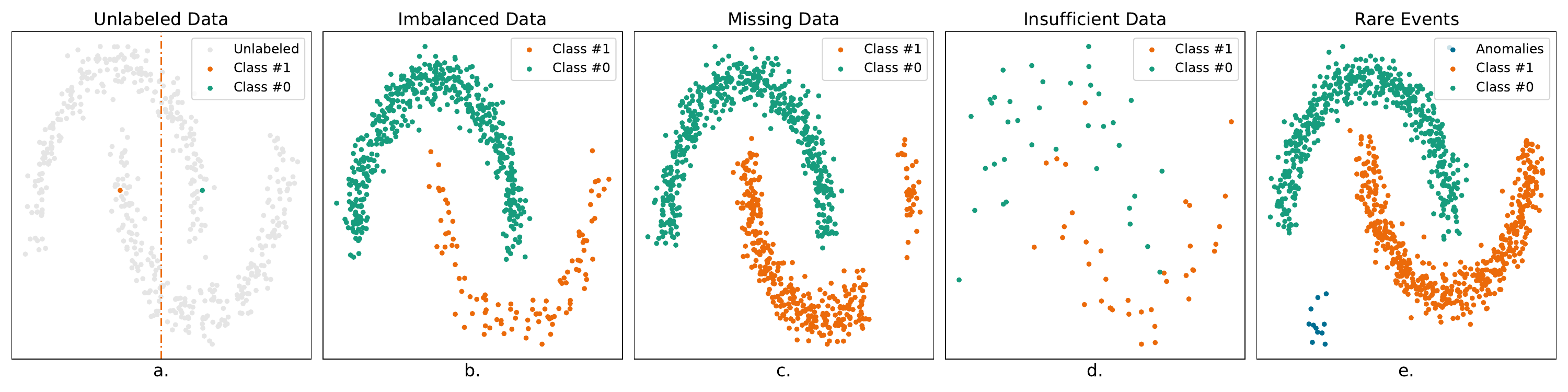}	
		\caption{Small data examples on the "half moons" synthetic dataset, which presents unique challenges: a) lacks annotations, b) exhibits class imbalances, c) contains missing data points, d) has a limited number of data points, and e) includes unseen or uncommon observations.}
		\label{fig:SmallData}
	\end{figure*}
	
	\subsection{Partly Labeled and Unlabeled Data}\label{ssec:PLandUD}
	A partially labeled or unlabeled data set is not necessarily small. However, only a small portion or even none of the data samples are labeled. Typically, this kind of data exhibits some small data properties, e.g., limited scope, weak relationality, and inflexible generation.
	
	Partly labeled or unlabeled data frequently occur in industrial applications: most machinery is equipped with all kinds of sensors and their readings can be logged. The entirety of such sensor readings may or may not contain useful information. Without labeling, it is hard to make sense of the data. Moreover, it is usually not possible to take advantage of data collected in other domains, either labeled or unlabeled, as the sensor readings are precise and restricted to the particular task at hand. Manual or semi-automatic analysis requires substantial effort involving organizational, technical, and human resources. 
	
	However, as implied by Fig.~\ref{fig:SmallData}-a, unlabeled data may be used for machine learning, particularly for unsupervised learning.
	
	\subsection{Imbalanced Data}\label{ssec:ID}
	A data set is said to be \emph{balanced} if all classes are (at least roughly) evenly represented. Formally, let
	\begin{equation}
	\alpha_i := \frac{N_i}{N_{\text{labeled}}}
	\qquad\text{with }i\in{1,2,\ldots K}
	\end{equation}
	be the quota of samples of class $i$. A data set is balanced if
	\begin{equation}
	\forall_i\in\{1,2,\ldots K\}: \alpha_i\approx\frac{1}{K}.
	\end{equation}
	Otherwise, the data set is called \emph{imbalanced}. Note that the definition is not strict. Particularly, we do not require all quota to be \emph{exactly} equal.
	
	The degree of imbalance (imbalance ratio IR) is measured differently, e.g., \cite{OIL17} simply use
	\begin{equation}
	I\!R = \frac{\max\nolimits_{i=1}^K\alpha_i}{\min\nolimits_{i=1}^K\alpha_i}.
	\end{equation}
	As the $\alpha_i$ are estimators of a discrete probability distribution, $\sum\nolimits_{i=1}^K \alpha_i =1$, it is natural to use the entropy as a measure of imbalance
	\begin{equation}
	H = - \sum\limits_{i=1}^K \alpha_i\log\alpha_i.
	\end{equation}
	In order to obtain comparable values for different numbers $K$ of classes, \cite{DKW09} normalize to the maximal entropy $H_0=-\log K$ and propose
	\begin{equation}
	\label{eqn:IR2}
	I\!R_{H} = \frac{H}{H_0}
	= \frac{\sum\nolimits_{i=1}^K \alpha_i\log\alpha_i}{\log K}
	\end{equation}
	as \emph{imbalance ratio}. $I\!R_{H}$ takes values in the interval $[0,1]$ where $0$ means totally imbalanced -- all samples from only one class -- and $1$ means perfectly balanced, i.e., $\alpha_1=\alpha_2=\ldots=1/K$.
	
	Fig.~\ref{fig:SmallData}-b shows a dataset with somewhat imbalanced data: $N_1=500$ and $N_2=100$
	samples. The imbalance ratio according to Eqn.~(\ref{eqn:IR2}) is $0.65$.
	
	Typically, imbalanced data occur in applications where large amounts of data are available, but only a few samples exhibit the desired traits. As machine learning algorithms usually rely on more or less balanced data, appropriate measures against data imbalance must be taken before machine learning. Otherwise, the learned models will likely show poor performance. 
	
	\subsection{Missing Data}\label{ssec:MD}
	Missing or incomplete data typically occur due to technical issues during data collection, such as, sensor failure, human error, or due to a flawed experimental or data collection procedure. Missing data reduce the representativeness of the samples, can introduce bias in the statistical analysis, and impedes matrix computations. In small data, the missing information cannot be reliably completed through a fusion of data of different types like in big data.
	
	\begin{figure}
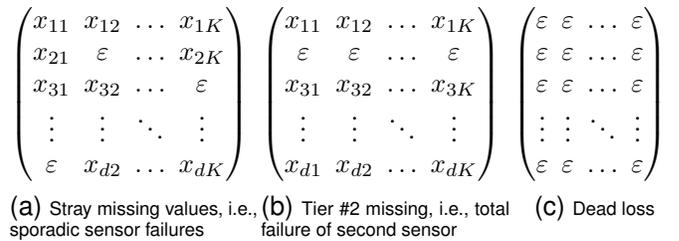

		\begingroup
		\arraycolsep2.5pt
		\centering
		\subfloat[\scriptsize{Stray missing values, i.e., sporadic sensor failures}]{%
			$\begin{pmatrix}
			x_{11} & x_{12} & \ldots & x_{1K} \\
			x_{21} & \varepsilon & \ldots & x_{2K} \\
			x_{31} & x_{32} & \ldots & \varepsilon \\
			\vdots & \vdots & \ddots & \vdots \\
			\varepsilon & x_{d2} & \ldots & x_{dK}
			\end{pmatrix}$
			\label{fig:fvsMiss-a}%
		}%
		\hfill\subfloat[\scriptsize{Tier \#2 missing, i.e., total failure of second sensor}]{%
			$\begin{pmatrix}
			x_{11} & x_{12} & \ldots & x_{1K} \\
			\varepsilon & \varepsilon & \ldots & \varepsilon \\
			x_{31} & x_{32} & \ldots & x_{3K} \\
			\vdots & \vdots & \ddots & \vdots \\
			x_{d1} & x_{d2} & \ldots & x_{dK}
			\end{pmatrix}$
			\label{fig:fvsMiss-b}%
		}%
		\hfill\subfloat[\scriptsize{Dead loss}]{%
			$\begin{pmatrix}
			\varepsilon & \varepsilon & \ldots & \varepsilon \\
			\varepsilon & \varepsilon & \ldots & \varepsilon \\
			\varepsilon & \varepsilon & \ldots & \varepsilon \\
			\vdots      & \vdots      & \ddots & \vdots      \\
			\varepsilon & \varepsilon & \ldots & \varepsilon
			\end{pmatrix}$
			\label{fig:fvsMiss-c}%
		}%
		\caption{Examples of missing data, denoted by symbol $\varepsilon$, in 
			sequences of $K$ numerical feature vectors, i.e., $x_{ij}\in\mathbb{C}$.
			\label{tab:fvsMiss}}%
		\label{fig:fvsMiss}
		\endgroup
	\end{figure} 
	
	We consider numerical features and extend the definition of $d$-dimensional real or complex feature vector sequences---i.e., $\mathcal{X} \subseteq \big(\mathbb{C}^d\big)^{\!+}$, see  Sect.~\ref{ssec:MLFandDS}---as follows:
	\begin{equation}
	\label{eqn:fvsMiss}
	\mathcal{X}_\varepsilon 
	\subseteq \big((\mathbb{C}\cup\{\varepsilon\})^d\big)^{\!+}\!\!.
	\end{equation}
	Here, $\varepsilon$ stands for an absent value -- in databases usually denoted by \TT{NIL}. Fig.~\ref{fig:fvsMiss} shows some possible cases of missingness. 
	
	There are two extreme cases 
	a dead loss sample as shown in Fig.~\ref{fig:fvsMiss}-c: We could have a dead loss sample (Eq. \ref{eq:dead_loss}) with a valid label $(e,y\neq\varepsilon)\in\mathcal{D}$ or even a dead loss sample without label $(e,\varepsilon)\in\mathcal{D}$ which has its index in the dataset determined either explicitly by the data acquisition protocol (e.g., the occurrence of an empty recordings file) or implicitly (for, e.g., samples with timestamps). 
	
	\begin{equation}
	e = \begin{pmatrix}
	\varepsilon & \ldots & \varepsilon \\
	\vdots      & \ddots & \vdots      \\
	\varepsilon & \ldots & \varepsilon
	\end{pmatrix}\in\mathcal{X}_\varepsilon
	\label{eq:dead_loss}
	\end{equation}
	
	Approaches to handle such cases are presented in Section \ref{ssec:AtMD}.
	
	Every data point has some likelihood of being missing, and according to the mechanisms of missingness, there are three types of missing data\cite{R76,LR19}: 
	
	\emph{Missing at random (MAR)}, for given observed data, the probability of missing data is independent of the unobserved data. However, it is related to some of the observed data, see Fig.~\ref{fig:fvsMiss}-a. 
	A common approach to MAR data is either to omit cases with any missing data on the variables of interest or to predict the values by other features in the data.
	For example, the older the car, it is more probably that the mileage will not be provided by the seller. However, the missingness of the mileage of the car can be predicted from its manufacturing year.
	
	\emph{Missing completely at random (MCAR)}, is the case where the probability that data are missing is independent of both observed and unobserved data. There is no pattern and no bias is introduced, and the missing data is a random subset of the data. 
	The analyses can be performed using only observations with complete data considering them a simple random sample of the entire data.
	In reality, data seldom exhibit characteristics of MCAR. For instance, if there are occasional missing values in the sensor readings because of a power outage, this would be MCAR, see Fig.~\ref{fig:fvsMiss}-b. However, if there is any pattern in the outages, it would be correlated with the sensors' readings, then the data is not MCAR.
	
	\emph{Missing not at random (MNAR)}, data that do not meet the description of MCAR nor MAR. Here the missingness is specifically related to the missing variable. Since none of the standard methods for dealing with missing data can be employed, advanced predictive or statistical modeling of the missing data is necessary to get an unbiased estimate of the parameters.
	For instance, if the temperature sensor fails whenever there is a condition that meets those out of the operational range (Fig.~\ref{fig:SmallData}-c) (too low or too high-temperature values), or it failed because of the power supply outage caused by the condition itself.
	
	\subsection{Insufficient Data}\label{ssec:InsDa}
	Data are called insufficient if their volume is very small and intrinsically limited due to expensive collection \cite{K09} or generation, or very low frequency of production. With insufficient data, machine learning algorithms cannot obtain proper models. 
	
	Fig.~\ref{fig:SmallData}-d gives an example of a dataset with 75 samples per class, where it is evident that it is difficult to determine the optimal boundary between the two classes.
	
	\subsection{Rare Events}\label{ssec:RE}
	Rare events, also known as anomalies and novelties, could be described as rare observations that deviate from the normal operations or usual behavior. They could result from data acquisition or data processing errors and natural variation or foreign influence in the system.
	
	Due to a large amount of ``normal'' data, detecting rare events or anomalies is usually related to big data and veracity. However, very few and rare observations contain valuable information of high interest for the intended task. Hence the anomalies themselves fit the definition of small data as given in the Sect.~\ref{ssec:DefSD}.
	In industrial and engineering applications, detection or prediction of such rare events is crucial, especially in predictive maintenance and quality assurance.
	
	The authors of \cite{CBK09} classify anomalies into three categories: 
	
	\emph{Global outliers or point anomalies} represent individual observations considered anomalous wrt. the rest of the data, usually due to data pollution by noise and outliers. Point anomalies are not distributed into dense regions or clusters, and the observations lie outside the boundaries of the normal regions.
	
	\emph{Contextual or conditional anomalies} are observations considered anomalous in a specific context (temporal or spatial) or specific conditions, only. Outside the given context, the observations would be considered normal.
	
	\emph{Collective anomalies} are a collection of observations that form clusters that are considered anomalous with respect to the entire data in the sense that there is no model for the cluster (see Fig.~\ref{fig:SmallData}-e). There are two typical cases: a) data are anomalies or b) the data points represent a new previous unseen class which is normal and should be included in the model. 
	
	Anomalies are dynamic. Emerging new types of observations make it difficult to obtain or define labels for such occurrences. In order to formally describe an anomaly detection problem, we first consider a set of two labels
	\begin{equation}\label{eq:aslabels}
	\mathcal{Y} := \{c_{\text{normal}}, c_{\text{anomaly}}\}.
	\end{equation}
	The training set for machine learning contains normal samples only
	\begin{equation}\label{eq:addataset}
	\mathcal{D}_{\text{train}} := \big\{(x,y\!=\!c_{\text{normal}})\big\}.
	\end{equation}
	Thus, we can learn a model for the $c_{\text{normal}}$ class only. An anomaly detector is to find out whether an unknown observation $x_\text{test}$ is normal or an anomaly. Since we have only one class model, the detection decision $y_\text{test}$ must be based on some kind of score $s(x)\in\mathbb{R}$ indicating how well the normal model fits the observation
	\begin{equation}
	\label{eq:anomalous}
	y_\text{test} 
	= \begin{cases}
	c_\text{anomaly} &: s(x_\text{test}) < \theta \\
	c_\text{normal}  &: \text{otherwise},
	\end{cases} 
	\end{equation}
	where $\theta\in\mathbb{R}$ denotes a decision threshold. The unknown observation $x_\text{test}$ is considered as anomaly if it scores lower than a defined threshold $\theta$.
	
	\section{Machine Learning with Small Data}\label{sec:MLwSD}
	This section presents a taxonomy of ML approaches and algorithms in the context of small data in industrial applications.
	We will refer to the definition given by T.M. Mitchel \cite{MT97} in Section~\ref{sec:Intro} where a machine learning problem is defined 
	as ``learning from experience ${E}$ concerning some class of tasks ${T}$ and performance
	measure ${P}$, if its performance at tasks in ${T}$, as measured
	by ${P}$, improves with experience ${E}$.''
	
	Experience ${E}$ can be defined by the domain and its representation as dataset $\mathcal{D}$, along with the learning paradigm. It could be considered a function of an ML approach and supervised information in the form of data $\mathcal{D}$, expert, and ``a priori'' knowledge.
	The type and nature of data $\mathcal{D}$ along with the task ${T}$ will determine a suitable approach for solving a particular problem.
	
	In the context of industrial applications, we take the example of product quality prediction. The task ${T}$ is to distinguish faulty products from good ones. The performance measure ${P}$ is detection accuracy, based on training experience ${E}$ (knowledge about what is good) gained by either learning on data or using ``a priori'' (domain) knowledge about the physical properties of good specimens.
	
	From here, having examples of only good specimens ${E}$ to detect bad ones ${T}$ defined the machine learning approach as an anomaly or novelty detection.  
	
	\subsection{Domain Representation}\label{ssec:DR}
	\begin{figure}[!t]
		\centering
		\includegraphics[width=0.65\linewidth]{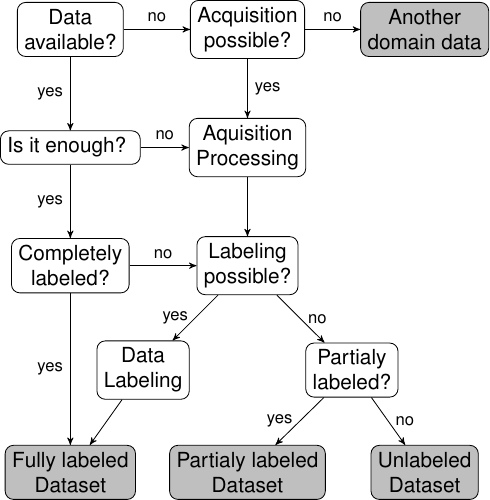}	
		\caption{Data collection process.}
		\label{fig:DC}
	\end{figure}
	
	The first and most important step in applying any machine learning approach is to ensure domain representative data ($\mathcal{D}$). Fig.~\ref{fig:DC} gives an overview of the requirements equally applied to existing data or data that should be collected.
	
	\begin{figure*}[!tb]
		\centering
		\includegraphics[width=0.85\linewidth]{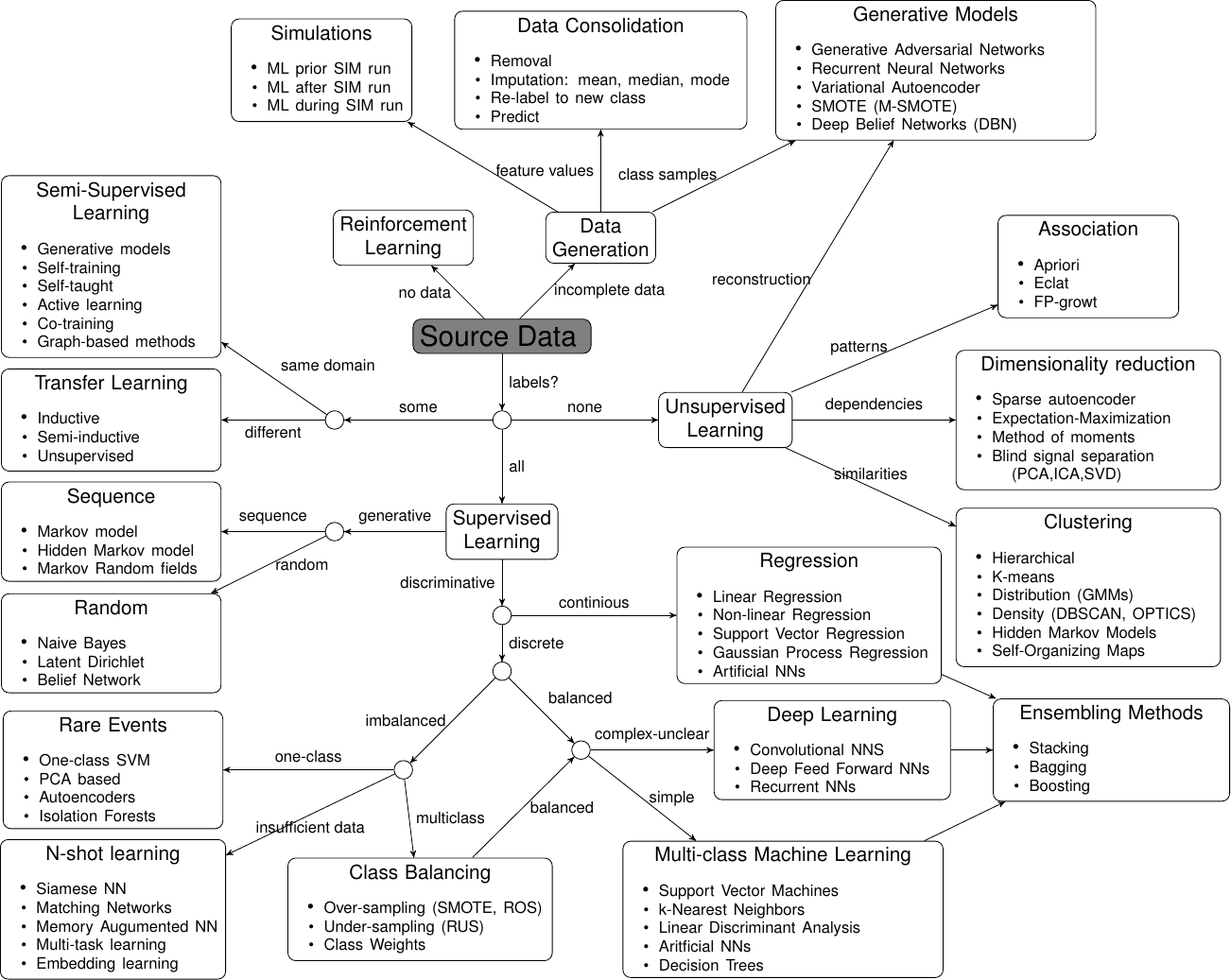}	
		\caption{Machine learning algorithms in context of small data.}
		\label{fig:TX}
	\end{figure*}
	
	The most important question is: do data already exists? If not, is their acquisition possible at all? Data collection might be technically and organizationally plausible, but it was never done before. Also, data could be unavailable because of infeasible and/or expensive acquisition, or available but very limited in terms of quantity and quality.
	
	In the case when data's existence is assumed, it might be that it was not observed before and could not be captured. Many such examples exist in industrial applications, e.g., in predictive maintenance and fault monitoring.
	
	The most extreme case is that there are no assumptions of the form and properties of the data, and only the task, context, or the domain of the application are known. Then using data collected in another domain or for another purpose could be employed in reinforcement \cite{SBRL18} or transfer learning \cite{ZQDXZZ21}, see Section~\ref{sssec:TrLrn}.
	
	Another important aspect is data annotation, where the collected data could differ regarding the labeling coverage as unlabeled, partially, and fully labeled. The availability, the size, and the annotation are the main factors determining the appropriate machine learning approach in the context of small data.
	
	\subsection{Taxonomy of Machine Learning}\label{ssec:MlTax}
	Fig.~\ref{fig:TX} presents a taxonomy inspired by similar ones as found in different studies and ML frameworks (such as \cite{GBC16,KPM21}). Its objective is to address the most popular approaches and algorithms relevant in the case of small data in industrial applications. 
	
	As already mentioned, the data $\mathcal{D}$ and the task $\mathcal{T}$ will help to make the right decisions and to select the most appropriate solution. Answering the requirements about the availability, completeness and the annotations of the data, determine one of the main paradigms in the machine learning: \textit{data generation}, \textit{supervised}, \textit{semi-supervised}, \textit{unsupervised}, \textit{transfer} and \textit{reinforcement learning}.
	
	Depending on the goal, other factors are also important for the optimal choice, \textit{discriminative} or \textit{generative} machine learning, \textit{discrete} or \textit{continuous} target variables, \textit{balanced} and \textit{imbalanced} data, complexity and so on.
	
	Discriminative models learn the conditional probability of the target given the observable variable, 
	while the generative models compute the joint probability distribution of the targets and the observations (distribution of the individual classes) \cite{J12}. 
	
	The target variables could be quantitative and categorical. Categorical variables are discrete and have a finite number of distinctive groups. The quantitative variables are numerical, discrete (finite number of values) or continuous (infinite number of values in a defined range). 
	
	In the case of when we have incomplete data, where target variable samples or feature values are missing, data generation and consolidation should be applied, where the resulting synthetic or surrogate data could still exhibit the limitations of small data.
	
	When the target variables are unknown, not defined, or not labeled, the appropriate approach depends on the criteria used to exploit the hidden and unknown patterns in the data: clustering by their similarities, discovering hidden dependencies for dimensionality reduction or reconstruction with generative models.
	
	Although it seems not apparent initially, even small data could be very complex and challenging to model with traditional ML approaches. Complexity could be expressed by a set of metrics (e.g., class ambiguity, boundary complexity, sample sparsity, and features dimensionality) \cite{BH06}, that quantify the data and indicate how complicated the pattern extraction and classification will be.
	
	\section{Approaches for Small Data}\label{sec:AtSD}
	In the following sections, we will focus on our taxonomy's intersecting points with the small data challenges as described in Section \ref{sec:ChSD}. Some notable approaches are also graphically presented for better understanding. 
	
	\begin{table}[!t]
		\centering
		\caption{Symbol Legend.}
		\label{tab:legend}
		\begin{tabular}{m{1.2cm}|c||m{1.2cm}|c} 
			\hline 
			\includegraphics[width=0.07\textwidth]{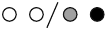} & unlabeled / labeled & \includegraphics[width=0.05\textwidth]{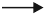} & sequence of events\\ 
			\hline 
			\includegraphics[width=0.03\textwidth]{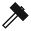} & construction & \includegraphics[width=0.03\textwidth]{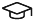} & learning\\ 
			\hline 
			\includegraphics[width=0.03\textwidth]{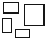} & features & \includegraphics[width=0.04\textwidth]{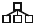} & classification\\ 
			\hline 
			\includegraphics[width=0.05\textwidth]{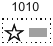} & embedding & \includegraphics[width=0.05\textwidth]{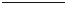} & relation\\ 
			\hline 
			\includegraphics[width=0.03\textwidth]{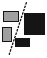} & models & \includegraphics[width=0.05\textwidth]{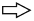} & usage\\ 
			\hline 
			\includegraphics[width=0.03\textwidth]{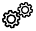} & transformations &
			\includegraphics[width=0.05\textwidth]{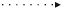} & feedback\\ 
			\hline 
			\includegraphics[width=0.05\textwidth]{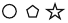} & feature - form & \includegraphics[width=0.05\textwidth]{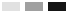} & feature - grade \\
			\hline
			\includegraphics[width=0.03\textwidth]{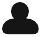} & expert & 
			\includegraphics[width=0.05\textwidth]{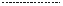} & boundary\\
			\hline
			\includegraphics[width=0.03\textwidth]{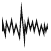} & noise & 
			\includegraphics[width=0.05\textwidth]{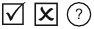} & real/fake/unknown\\
			\hline
		\end{tabular}
	\end{table}
	
	The legend of the used symbols is given in Table~\ref{tab:legend}. The input data points could be \emph{unlabeled} and \emph{labeled}. They can have different attributes such as e.g., one geometrical form, another grade of brightness. Hidden dependencies and \emph{relations} could be discovered automatically or by human \emph{experts}.
	The data could be \emph{transformed} or converted into another \emph{feature} space and their \emph{embeddings} can be estimated. 
	The \emph{learned model} knows about the features and \emph{classification boundary} between the classes. The flow of the symbols in the graphs is from left to right and from top to bottom.
	
	\subsection{Approaches to Unlabeled Data}\label{ssec:AtUD}
	This section discusses different approaches that could be employed to overcome the challenges in machine learning with unlabeled data as already introduced in Section~\ref{ssec:PLandUD}. 
	
	The first and most obvious solution is to label the whole amount of the available data by experts, by crowd-sourcing annotators, by synthetic labeling, or by data programming (programmatic creation of training sets) \cite{RDSWSR2017}. Alternatively, human experts can smartly select a small subset of unlabeled data to be annotated and use it as training data to create a machine learning model. This approach is also known as active learning \cite{S09}.
	However, depending on the availability of human, time, and financial resources, labeling might not be possible or feasible even in the case of small data.
	The main strategies can be grouped depending on how much of the available data is labeled (some or none) and whether the labeled and unlabeled data are of the same type:
	\begin{itemize}
		\item no labeled data at all: unsupervised learning \cite{HSP99}
		\item small amount of labeled data:
		\begin{itemize}
			\item semi-supervised learning \cite{CSZ09,VJH20}, where the labeled and unlabeled data are of the same type and 
			\item transfer learning \cite{PY10,WKW16,ZQDXZZ21}, where the labeled and unlabeled data are from a different type.
		\end{itemize}	
	\end{itemize}
	
	All the mentioned approaches share standard techniques that could be equally used with small labeled datasets and unlabeled data. 
	
	\begin{figure}[!t]
		\centering
		\includegraphics[width=0.8\linewidth]{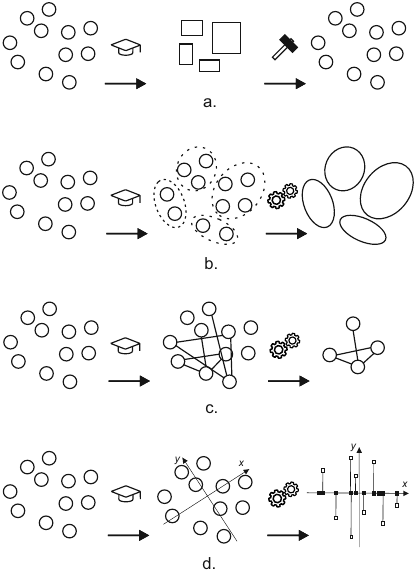}	
		\caption{Unsupervised learning: a-reconstruction (generative models), b-similarities (clustering), c-patterns (associations), d-dependencies (dimensionality reduction).}
		\label{fig:UL_cluster}
	\end{figure}
	
	\subsubsection{Unsupervised Learning}\label{sssec:unsup-l}
	In unsupervised machine learning, algorithms try to identify clusters, to discover unknown classes, or to point out outliers:
	\begin{itemize}
		\item Generative models Fig.~\ref{fig:UL_cluster}-a: auto-encoders \cite{B12}, particularly stacked auto-encoders were successfully applied to predict material defects with small datasets \cite{FZD19}, deep belief nets \cite{MDH09}, generative adversarial networks \cite{GPMXWOCB14};
		\item Clustering \cite{GCB04} Fig.~\ref{fig:UL_cluster}-b: hierarchical, $k$-means, mixture models, density based, hidden Markov models, self-organizing maps \cite{K90};	
		\item Association rule mining (ARM) Fig.~\ref{fig:UL_cluster}-c: discovering relationships between different entities in non-numeric datasets when the variable of interest is not known \cite{KK06}. The basic approaches are the Apriori \cite{AIS93}, the Eclat \cite{Z00} and the FP-growth \cite{JJY00} algorithm. ARM approaches were used in many industrial applications and manufacturing \cite{DB20}, such as in bearing defects diagnosis in \cite{BZWDW16}, fault diagnosis of power transformers \cite{YTS09};
		\item Dimensionality reduction Fig.~\ref{fig:UL_cluster}-d: expectation–maximization (EM) algorithm \cite{DLR77}, method of moments, blind signal separation such as principal component analysis (PCA) \cite{H02}, independent component analysis (ICA) \cite{HO00}, singular value decomposition (SVD) \cite{MHW03}.
	\end{itemize}
	
	\subsubsection{Semi-supervised Learning}\label{sssec:semi-l}
	Here it is assumed that the unlabeled and labeled data have the same distribution and that the data can be annotated using the same labels as the classification task. From here, heuristics can be used in order to annotate the unlabeled data based on some similarity with the labeled one. Additionally, the labeled data could be used directly in training \cite{BD18} for weight initialization, feature learning, and compressed data representation.
	
	\begin{figure}[!t]
		\centering
		\includegraphics[width=0.8\linewidth]{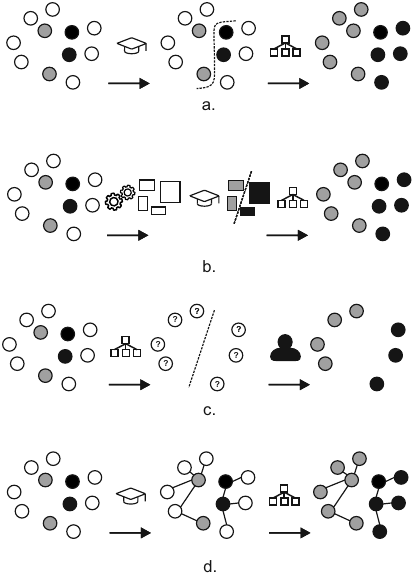}	
		\caption{Semi-supervised learning: a- self training, b- self taught, c- active learning, d- graph methods.}
		\label{fig:ss1}
	\end{figure}
	
\begin{figure}[H]
		\centering
		\includegraphics[width=0.8\linewidth]{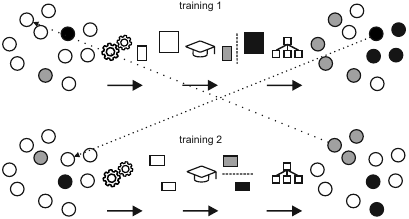}	
		\caption{Co-training: use complementary information to train model and label data.}
		\label{fig:CoTr}
	\end{figure}

	The following methods are commonly used for semi-supervised learning \cite{Z05, ZG09}: 
	\begin{itemize}
		\item Self-training, Fig.~\ref{fig:ss1}-a: train a classifier, apply it to unlabeled data, and add the predicted labels into the training set;
		\item Self-taught, \cite{RBLPN07}, usage of unlabeled data to construct a new representation, express the labeled data in the new representation, and use existing classification methods on this new representation space, Fig.~\ref{fig:ss1}-b, e.g., using unlabeled data with autoencoders to learn features which can be used for supervised learning with the labeled data;
		\item Active learning, Fig.~\ref{fig:ss1}-c: the ML algorithm query unlabeled data instances to be labeled by an oracle (e.g., a human annotator) \cite{S09};
		\item graph-based methods, Fig.~\ref{fig:ss1}-d: Label propagation, similar data points have similar labels, the information "propagates" from labeled data points \cite{ZX05}.
		\item Co-training, Fig.~\ref{fig:CoTr}: provide different and complementary information about the data to the ML algorithm \cite{NWXCZY21};
		\item Generative models: generative adversarial networks GANs \cite{GPMXWOCB14}, autoregressive models and probabilistic deep generative models like variational auto-encoder \cite{KW13};
	\end{itemize}
	
	\subsubsection{Transfer Learning}\label{sssec:TrLrn}
	Transfer learning acquires knowledge by training on labeled data from one task and utilizing it to solve different, yet, related tasks. The learning process is much faster, more accurate, and needs less training data than creating model from scratch.
	Other terms often related to transfer learning may be knowledge-transfer, inductive and semi-inductive-transfer, meta-learning, incremental learning, etc. 
	
	For a better explanation, we will stick to terms used in \cite{PY10, WKW16, ZQDXZZ21}. First, we define a source domain $\mathcal{D}_{s}$, containing a feature space, the marginal probability distributions of its elements. The corresponding source task $\mathcal{T}_{s}$ contains a label space and predictive objective functions $f_{s}(\mathbf{x})$ which are trained on the source dataset and the source labels. The feature space contains $N_s$ feature vectors (patterns), which must be learned.
	
	Next we define a target domain $\mathcal{D}_{t}$ and the corresponding target task $\mathcal{T}_{t}$ also containing a label space and objective functions $f_{t}(\mathbf{x})$. Usually the dimension or size of the feature space in the target domain ($N_t$) is smaller then the source domain ($1 < N_t \ll N_s$) representing the ``real'' small data problem.
	
\begin{figure}[!h]
		\centering
		\includegraphics[width=0.8\linewidth]{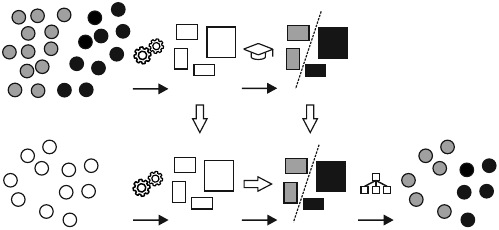}	
		\caption{Transfer learning: use abundant data from other domain to discover features and train a model that can be used for the target domain.}
		\label{fig:trl}
	\end{figure}

	The main objective of transfer learning is to derive a predictive objective function $f_{t}$ for the target task $\mathcal{T}_{t}$ using information derived from $\mathcal{D}_{s}$ and $\mathcal{T}_{s}$, see Fig.~\ref{fig:trl}. Different sub types of transfer learning can be defined depending on the similarity of source and a target domain and task data. 
	
	In this paper we can only give a brief overview. We refer to the literature for details. Hence we only use some common sub-types and point out specific requirements. Categorization wrt. to labels of data set:
	\begin{itemize}
		\item{Inductive transfer learning: requires some labeled data for target domain $\mathcal{D}_{t}$,}
		\item{Semi-inductive transfer learning: no or little labeled data in target domain $\mathcal{D}_{t}$, and}
		\item{Unsupervised transfer learning: no labeled data at all in $\mathcal{D}_{t}$ and $\mathcal{D}_{s}$.}
	\end{itemize}
	
	The delimitation of the types of transfer learning differs in the literature, see detailed overview in \cite{WKW16,ZQDXZZ21}. As the last part, it is to explain what kind of knowledge is transferred and how. 
	
	According to \cite{ZQDXZZ21} four general knowledge transfer categories are dominant distinguished by transfer learning method.
	
	Category \textit{instance-based transfer learning} is a common method to pick parts of the source datasets and reuse them by re-weighting and feeding them to the ``usual'' learning process using any classifier for the target domain. The learning success is tested on the target dataset. Thus this method is only applicable if some target data is available.  
	
	The second category is \textit{feature-based learning}. This method aims to extract meaningful structures between source and target domains to find a common feature space that has predictive qualities for both domains.
	
	The third category is \textit{parameter-based learning}. This method is based on an underlying assumption that source and target tasks share the same parameters. Hence parameters must be encoded.
	
	Fourth category is \textit{relational based learning}. This method is based on extracting similar relations in target and source domains\cite{SK15}.     
	There are quite a few studies related to transfer-learning for industrial and engineering applications, but some transfer learning cases in aerospace can be found within \cite{MSGOG17}.
	
	\subsection{Approaches to Imbalanced Data}\label{ssec:AtID}
	As defined in Section~\ref{ssec:ID}, imbalanced data problems occur when one class has a significantly smaller number of samples than others, for instance, in binary classification tasks, the minority vs. majority class \cite{HYSM17}.
	This can be extended to multi-class classification scenarios \cite{WY12}. The notion of class imbalance is also called \emph{class rarity} or \emph{skewed data} \cite{JK19}.
	
	In many real-world applications \cite{Ja00}, the minority class is usually of greater interest since useful information belongs to this class, 
	e.g. medical diagnostics \cite{MG06, LLH10, PMA11, WYC21}, detection of banking fraud \cite{WLCOC13}, network intrusion \cite{CCS06} and oil spills \cite{OCW17}.
	
	The core issue of imbalanced learning algorithms is that the minority class is often to be miss-classified because prior information of the majority class has more influence during the training and consequently on the prediction \cite{JK19}.
	
	According to \cite{K16, JK19}, there are mainly three strategies to deal with class imbalance: data-level methods, algorithm-level techniques, and hybrid approaches. Here, we only provide main ideas on each topic.
	
	For detailed explanation and a complete list of references, we refer to \cite{JS02, W04, HG09, K16, HYSM17, JK19}.
	
	\subsubsection{Data-level Methods}\label{sssec:DlM}
	These methods' main idea is to try balancing out between majority and minority classes by making use of sampling methods \cite{HG09}.
	\begin{figure}[!h]
		\centering
		\includegraphics[width=0.9\linewidth]{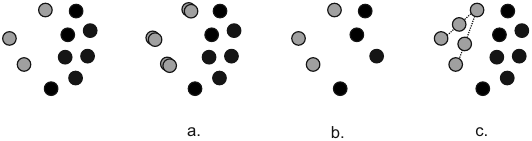}	
		\caption{Data level methods: a- ROS, b- RUS, c- SMOTE.}
		\label{fig:ID_D}
	\end{figure}
	
	This includes over-sampling minority classes \cite{CLWN13} and under-sampling majority classes, 
	e.g., synthetic minority over-sampling technique (SMOTE) \cite{CBHK02}, 
	random under-sampling (RUS) and random over-sampling (ROS) methods \cite{VKN07}, as presented on Fig.~\ref{fig:ID_D}.
	Moreover, several deep learning-based data-level methods have been proposed:
	To achieve class balance, ROS and RUS have been applied to minority and majority class, respectively \cite{BMM18}. 
	Also, a dynamic sampling technique based on two-phase learning with ROS and RUS \cite{LPK16} and convolutional neural network (CNN) has been proposed \cite{PTMTKGDALCS18}.
	
	\subsubsection{Algorithm-level Methods}\label{sssec:AlM}
	While the aforementioned data-level methods change the training data distribution directly, the primary focus of algorithm-level methods is to modify existing algorithms so that more emphasis is placed on minority classes \cite{LLW02, AKJ04, WC05}.
	
	\begin{figure}[!h]
		\centering
		\includegraphics[width=0.8\linewidth]{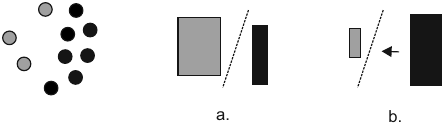}	
		\caption{Algorithm level methods: a- class weights, b- decision threshold.}
		\label{fig:id_a}
	\end{figure}
	
	This can be achieved by introducing weights to classes (Fig.~\ref{fig:id_a}-a) via a cost matrix \cite{JK19}, e.g., a more weight for each minority class,
	which is called cost-sensitive learning \cite{FGGPKH18}.
	
	However, it is difficult to find an optimal cost matrix since this process usually requires domain-specific expert knowledge and is not feasible in case of large data set or large number of features \cite{JK19}.
	
	One of the biggest challenges in cost-sensitive learning is the assignment of an effective cost matrix. The cost matrix can be defined empirically, based on past experiences, or a domain expert with knowledge of the problem can define them. Alternatively, the false negative cost can be set to a fixed value while the false positive cost is varied, using a validation set to identify the ideal cost matrix. The latter has the advantage of exploring a range of costs, but can be expensive and even impractical if the size of the data set or number of features is too large.
	
	Another approach is to adjust a decision threshold (Fig.~\ref{fig:id_a}-b) in test time in such a way that bias moves from majority classes towards minority classes \cite{CTMAYC06, KW15}.
	
	Moreover, CNN-based methods have been proposed too, e.g., using a different loss function \cite{LGGHD17} or by moving the decision threshold \cite{BMM18}.
	
	\subsubsection{Hybrid Methods}\label{sssec:HM}
	For imbalanced data problems, hybrid methods refer to approaches that combine data-level and algorithm-level methods \cite{WLCC12}.
	
	As an example, one can apply cost-sensitive learning  
	after ensemble methods with data sampling \cite{HG09, JK19}.
	According to \cite{DGZ18}, the classical hybrid methods can usually be applied to small scale problems, characterized by non-extreme imbalance ratio and problem-specific hand-crafted low dimensional features.
	
	Recently, CNN based methods have also been proposed, which include:
	large margin local embedding (LMLE) by combining triple header hinge loss function and quintuplet sampling \cite{HLLT16} and deep over-sampling (DOS) 
	using micro-cluster loss, $k$-nearest neighbors ($k$-NN) and 
	over-sampling of a minority class \cite{AH17}.
	
	\subsection{Approaches to Missing Data}
	\label{ssec:AtMD}
	Apart from the types of missingness described in Section~\ref{ssec:MD}, missing data can also be interpreted as missing feature values or class samples. According to that, standard techniques dealing with missing data can be divided into three categories. 
	
	The first category is related to consolidate and prepare data for machine learning by removing data with missing values, imputation, and in the case of missing labels, to re-label, and predict labels \cite{E10}.
	The second category encompasses methods for data generation with generative models based on deep neural networks: generative adversarial networks (GANs) \cite{GPMXWOCB14}, variational autoencoders (VAEs), and Deep Belief Networks (DBN), including synthetic minority oversampling technique SMOTE. 
	The third group comprises of simulation-driven approaches.
	
	\subsubsection{Data Consolidation}\label{sssec:DC}
	In this context, missing data can be handled in three different ways. The first and most straightforward approach is to accept the data as it is.
	The second one is to delete the data with missing values from the dataset. That could be done by, either list-wise deletion, where the observations are removed when even one value of the features is missing, or by pair-wise deletion where it is attempted to use all available data and omit cases on an analysis by analysis basis, e.g., correlational analysis \cite{PE04}. Removing data reduces the amount of valuable information since many data analysis techniques, including machine learning, are sensitive to missing data. The third approach is the replacement of the missing values by single or multiple imputation and interpolation. Commonly used imputation techniques are mean, median, mode and regression imputation, Maximum Likelihood Estimation (MLE), and expectation–maximization (EM) algorithm. Interpolation creates new values for the missing ones by using the nearest feature value. A comprehensive survey of the current methodology for handling missing data is given in \cite{LR19} and missing data techniques in pattern recognition are reviewed in \cite{GSF10}.
	
	\subsubsection{Generative Models}\label{sssec:GM}
	In this approach, a generative model is trained from an incomplete dataset. The generative model is then used to generate plausible replacements for the missing values.
	The models represent the data space, and observations can be sampled by a random variable from the latent space. The deep generative models widely used for missing data generation are generative adversarial networks (GANs), variational autoencoders (VAEs), Deep Belief Networks (DBNs), and Recurrent Neural Networks (RNNs). 
	
\begin{figure}[!h]
		\centering
		\includegraphics[width=0.8\linewidth]{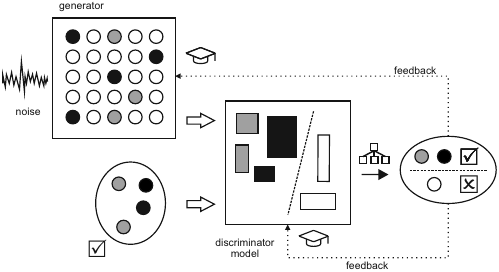}	
		\caption{Generative Adversarial Networks: a discriminator is deciding whether randomly sampled data from the domain space can be considered as real or fake.}
		\label{fig:GAN}
	\end{figure}

	\paragraph{Generative Adversarial Networks}
	GAN \cite{GPMXWOCB14} is a deep generative model which is capable of creating data with the same statistical properties as the available training data. The generative model produces a sample from the model, and an adversary (discriminative model) learns to distinguish the sample as belonging to the model or the data distribution.
	The basic idea could be explained in the context of a non-cooperative zero-sum game in game theory \cite{M97}.
	The two models are competing and improving their methods until the point where the generated samples are indistinguishable from the actual data samples (Fig.~\ref{fig:GAN}).
	
	An illustrative example is where a generator is trying to create images of counterfeit banknotes and a discriminator is trying to recognize whether images are fake or real.
	The generator's role is to create an image from the sample of input noise, e.g., noise using a normal or uniform distribution.
	The discriminator is determining the probability of how close the generated sample is to the real one. This information is then fed back into the generator as guidance to mimic essential features of the authentic images during the training.
	
	The network type is not restricted. A straightforward implementation consists of multlayer perceptrons for the generator and discriminator.
	In many applications, the generator is de-convolutional, and the discriminator a convolutional neural network. 
	The training is performed through back-propagation for both networks where they play a two-player minimax game to find the saddle point instead of the local optimum.
	The loss function of GAN is based on a cross-entropy such that the following odds are maximized:
	the created sample is classified as fake, and the reference one is recognized as real.
	The training procedure is performed in an alternating way utilizing the gradient descent (GD): When the generator's parameters are fixed, a single iteration of gradient descent is carried out on the discriminator. Then, another gradient descent is performed on the generator with the fixed parameters of the discriminator.
	When the optimum is reached, the generated samples are indistinguishable from real data, and the discriminator may then be discarded.
	Based on game theory \cite{M97}, the convergence of GAN can be achieved when the generator and the discriminator reach a so-called Nash equilibrium.
	
	The training of GAN is, however, highly challenging \cite{AB17} due to instability, non-convergence, mode collapse, or vanishing gradient \cite{H91}. However, the learning is performing well when the right architecture and the hyperparameters are selected.
	
	To deal with vanishing gradient and to improve the stability of GAN, Wasserstein GAN (WGAN) \cite{ACB17} has been proposed by making use of Wasserstein metric \cite{V69}, another approach has also been proposed in the context of semi-supervised learning \cite{SGZCRC16}.
	
	In recent years, the GAN moved from an exotic technique to mainstream, and the application area of GAN is extending rapidly.
	Major area of applications are in applied computer-vision, but there are increasing numbers in other areas, e.g.
	missing data imputation \cite{YJV18,LCZX18},
	image in-painting \cite{PKDDE16},
	image style transfer \cite{GEB16},
	3D object generation \cite{WZXFT16},
	realistic image synthesis \cite{RMC16, KALL18},
	object detection \cite{LLWXFY17},
	high resolution image blending \cite{WZZH17},
	visual surface inspection \cite{ZZCW18}, 
	anomaly detection in medical images \cite{SSWSL17} and industrial time series \cite{JHZHC19}, unsupervised fault detection \cite{SpBo18},
	generating molecules for drug discovery \cite{B17},
	synthetic medical patient records \cite{CBMDSS17}, 
	music generation \cite{ES02b, YCY17},
	cartography \cite{KGR19}, and
	astrophysics and cosmology \cite{SZZFS17, MBBLAK19}.
	
	\paragraph{Variational Autoencoders (VAE)}
	Autoencoders (AE) are deep neural networks that are primary used for learning representative features and for dimensionality reduction. 
	The basic training objective is to learn how to reconstruct the network's input at the output \cite{GBC16}. It is essentially a supervised learning task, although the data set is not required to be labeled.
	
	The architecture contains hidden layer with few neurons--called a bottleneck layer--which prevents the network from simply passing the input to the output. The layer effectively compresses the input into a code.
	The network consists of two parts, an encoder that transforms the input data to the code and the decoder which uses that representation to approximately reconstruct the input data as well as possible. It can be thought of as a standard feed-forward or convolutional network trained with mini-batch gradient descent and backpropagation, aiming to minimize the reconstruction error.
	There are variants of autoencoders which try to prevent learning the simple identity function and to improve the ability to capture the salient information. AEs can be regularized, sparse, de-noising, and contractive.
	
	Variational autoencoders are generative variant of autoencoders \cite{KW13}. The learning algorithm differs from ordinary autoencoders. VAE employ probabilistic mechanisms for describing an observation in the space of latent variables, where the encoder describes the probability distribution for each latent variable instead of providing a single value for each encoding dimension. 
	By sampling from the latent space, the VAE decoder part can generate new data similar to that which was used during training.
	Compared with GAN they are easier to train but provide worse results due to the MSE based loss function. Further information and insights about VAEs and their extensions are presented in \cite{D16}, and \cite{KW19}.
	Variational autoencoders were successfully used for missing data imputation in simulated milling circuit dataset \cite{MKA18}, traffic estimation \cite{BVMS19}, sensor data \cite{JTSP17} and soft sensors frameworks in industry \cite{XMHCH19}.
	
	\paragraph{Deep Belief Networks}
	Deep Belief Networks (DBN) \cite{HS06,H09} are probabilistic generative models consisting of multiple layers of hidden stochastic binary units with weighted connections.
	
	DBNs are hybrid graphical models and can be defined as a stack of simpler networks like RBMs (Restricted Boltzman machine) or autoencoders. They can be trained in an unsupervised learning task to discover deep hierarchical representations and reconstruct their inputs.
	The top layers form associative memory by undirected symmetric connections. The lower layers have directed acyclic connections to convert the associative memory to observable variables; the lowest layer states represent the data vector.
	
	DBNs are trained by an efficient layer-wise greedy learning using the Contrastive Divergence (CD) algorithm. The first layer models the data as a visible layer, then the learned representation is used as input data for the second layer. The process iterates for the subsequent layers until the last hidden layer is achieved, each time propagating the activation of the learned features upward and learn their higher-level representations. 
	Discriminative fine-tuning of a trained DBN can be done by adding a final layer converting the learned representations into desired supervised predictions and back-propagating error derivatives.
	The trained DBM can be used directly as a generative model by sampling from the top two hidden layers and then generate a sample from the model by drawing from the visible units in a single pass of forward sampling through the rest of the model.
	They are used for generating and recognizing images, video, and motion-capture data, for missing data imputation \cite{CLJXC15, ZZG14}, image restoration \cite{NTA13}, data recovery in sensor networks \cite{DCZ19}. 
	They are also successfully applied for classification in many other areas, e.g.
	fault detection \cite{TKZCYL18}, 
	vehicle detection \cite{WCC14}, 
	bearing fault diagnosis \cite{SJZL17}, 
	gearbox fault diagnosis \cite{CLS15}, 
	time-series forecasting \cite{KKKO14}, 
	structural health diagnosis \cite{TWW12}, etc.
	
	\paragraph{Recurrent Neural Networks}
	
	RNNs are artificial neural networks \cite{RHW88} which have at least one multiple fixed activation function unit (recurrent units) with a hidden state which signifies the past knowledge in a given time step \cite{H07, LBE15}.
	
	For the input sequence, an element at a time step and the hidden state information are fed into the recurrent unit, which updates the hidden state and returns an output, effectively memorizing the previous inputs and states. The process continues until the end of the sequence, when the RNN outputs the prediction.
	
	This property provides RNNs the capability of modeling long-term dependencies. 
	They are particularly useful for handling sequential data, such as audio or video signals, as well as stock market data. RNNs are employed in various tasks, including machine translation, time-series data analysis like automatic speech recognition \cite{DYDA12}, natural language processing \cite{YKYS17}, and video classification \cite{SZ14}.
	
	RNNs can be used also as a generative model 
	e.g., text generation \cite{SMH11}, image generation \cite{GDGRW15} and generation of complex sequences with long-range structure, such as text or online handwriting \cite{G13}.
	
	The main issues in training RNNs are the difficulties with the Gradient Descent optimization \cite{BBP13, PMB13}, the vanishing gradient or exploding gradient prevents learning of long data sequences.
	
	To overcome the vanishing gradient problems, gating-based architectures are proposed, such as, long short-term memory (LSTM) in \cite{HS97}, and later gated recurrent unit (GRU) in \cite{CMGBBSB14}. 
	The LSTM is composed of a cell, input, output, and forget gates. The memory cell keeps values over an arbitrary period, and the gates can control the inflows and outflows of information over time \cite{GSKSS17}. The GRU has less parameters, and it lacks the output gate compared to LSTMs; it achieves a performance comparable to LSTM on long sequences and even better performance on smaller datasets \cite{YYZY20}.  
	Variants of the basic RNN architecture are the bidirectional recurrent neural networks (BRNNs) \cite{SP97}, here the output layer get the input from two directions, the past and the future states in the same time.
	
	The application area of RNNs is quite extensive. In the form of generative models they were used e.g., for handling missing data in sequential series \cite{G13}, clinical time series \cite{LKW16}, and medical datasets \cite{YZvDS17}.
	
	\subsubsection{Simulation-driven Approaches}\label{sssec:S-dA}
	Simulation refers to, in general, an approach that uses computer programs to perform calculations based on mathematical models that represent real-world physical systems. It is employed when the mathematical model describing the physical system  too complex that it cannot be solved using traditional analytical methods.
	
	In this context, it is crucial to understand the properties of the system itself, including its underlying processes and the mathematical model in simulation-driven approaches for making decisions.
	That makes the difference between simulation-driven and the ML approaches that do not impose such specific requirements, which are, therefore called \emph{data-centric} or \emph{data-driven} approaches.
	
	There are three main categories of simulation-driven approaches: 
	
	\emph{Equation-based modeling (EBM)} \cite{F95, PSR98} which is typically described by ordinary differential equations (ODEs) or partial differential equations (PDEs) and thereby represents physical process well.
	
	\emph{Agent-based modeling (ABM)} \cite{EA96, W99, B02} which is well-suited for assessing the influence on a system described by actions and interactions of collective entities, i.e., agents.
	
	\emph{Hybrid modeling (HM)} makes use of both EBM and ABM, see e.g., \cite{W98, CKL15, MLG18}.
	
	Moreover, to integrate simulation models and ML, one may think of three scenarios: 
	
	\begin{enumerate}
		\item applying ML before a simulation run,
		\item applying ML after a simulation run, and
		\item applying ML within a simulation run.
	\end{enumerate}	
	In case (1), ML algorithms prepare input data for a simulation run \cite{EM18}.
	
	If the simulation models were agent-based models (ABM), the ML approach, e.g., a data-driven decision-making algorithm, must be designed so that the concept of collection of autonomous decision-making entities (agents) can be applied.
	
	In case (2), the simulation's output is fed directly into an ML algorithm, and one can find an application of this case in autonomous driving.
	Since training ML models for self-driving cars requires a large amount of data, the models are trained not entirely on real data but also on generated data from simulators, e.g., on real data from one city but on generated data for other cities that have similar characteristics as the real one \cite{DRCLK17}.
	
	Case (3) is very suitable for reinforcement learning (RL) \cite{G09}, e.g., finding an optimal path within a simulation setup. That means that the feedback of trying possible scenarios based on the simulation can directly be used during the ML models training, which is in contrast to cases (1) and (2).
	
	Simulation has played a significant role in science and engineering, and recent advances of ML push the boundaries of the field further \cite{vRMSBG20}.
	Hence, one can find numerous applications in various domains:
	network traffic control in telecommunication \cite{F89},
	physical systems modeling using ML and finite element method (FEM) \cite{KK18},
	optic flow estimation \cite{DFIHHGSCB15, IMSKDB17},
	fluid simulation using CNN-based generative model \cite{KATKGS19},
	reservoir simulation in shale gas production \cite{KME15},
	an agent-based operational simulation for unmanned aerial vehicles
	conducting maritime search-and-rescue missions in aerospace design \cite{SSF12},
	simulation-based kernel method for prediction modeling 
	using complex synthetic systems in bio-informatics \cite{DPWKSC19},
	DL approach to estimate stress distribution \cite{LLMS18},
	DL method for Reynolds-averaged Navier-Stokes simulations \cite{TWMMPH18},
	DL-based prediction model for cosmological structure formation \cite{HLFHRCP19},
	DL-based quantization simulation \cite{LPWK17},
	image generation method using simulated and unsupervised learning 
	by an adversarial network \cite{SPTSWW17},
	mechanical fault detection using simulation-trained classifier \cite{SFN18}.
	
	\subsection{Approaches to Insufficient Data}\label{ssec:AtLD}
	In the case of insufficient data, the dataset is complete and fully labeled. However, the boundaries between the target categories are ill-defined due to a small number of observations, in consequence the ML algorithms will not generalize well for unseen data. 
	
	In certain instances, the data exhibits an imbalance, and certain categories may be quite rare, with only a few observations available. Approaches to insufficient data can also apply to cases of imbalanced data (see Section \ref{ssec:AtID}). In any case of insufficient data, traditional ML approaches will fail to provide reliable and useful models. Possible approaches that could take advantage of limited supervised information and prior knowledge for rapid generalization are  N-shot learning (NSL) and transfer learning.
	
	When a sufficient number of observations are available to employ standard algorithms, addressing over-fitting requires the introduction of additional observations. This can be achieved through techniques such as data generation, augmentation, or the utilization of surrogate data.
	
	\subsubsection{N-shot Learning}\label{sssec:NSL}
\begin{figure}[htb]
		\centering
		\includegraphics[width=0.85\linewidth]{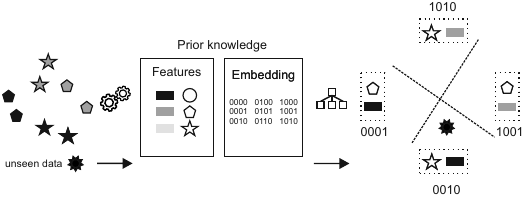}	
		\caption{N-shot learning: using prior knowledge about the data and the N samples per class, a boundary is found which can classify unseen data.}
		\label{fig:N_shot}
	\end{figure}

	N-shot learning approach aims to solve the problem of classification when only $N$ samples per class are available.
	The training data represents a support set $S$ with $N$ labeled samples of each class $K$, in total $N \times K$ observations, where objective is to classify a query set $Q$ of samples which lie in one of the classes Fig.~\ref{fig:N_shot}.
	
	When there is only one example per class, the support set $S$ has $N=K$ observations, and it is known as one-shot learning (OSL), or when there are few (less than five) then as few-shot learning (FSL)\cite{WYKN20}. Zero-shot learning (ZSL) is when the objective is to classify unseen classes without any observation available \cite{XLSA18, FXJXSG18} only by having a high level of descriptions \cite{GSB17, KCF18, LADOPB18}.
	One should note that humans have this type of learning ability \cite{LST15}, e.g., children can identify various different breeds of dogs even if they have encountered only a few of them. 
	However, the traditional machine learning and particularly deep neural networks (DNNs) based learning algorithms will fail in this problem since they typically require a larger amount of training data to achieve a certain level of accuracy \cite{KIH12, LBH15, SZ15, HZRS16, SBBWL16}.
	As discussed in \cite{WY19}, difficulties of OSL and FSL can be explained in terms of empirical risk minimization \cite{V91} by considering the interplay between approximation error and estimation error which depends on the sample size based on statistical learning theory \cite{Va95, HTF09}.
	Hence, several methods have been proposed by exploiting prior knowledge to deal with the unreliability as mentioned above of the empirical risk minimizer concerning data, model and algorithm.
	Here, we only tackled the basic ideas of each aspect, and for more details, we refer to \cite{WYKN20} and the references therein:
	
	\paragraph{Data-level Methods}\label{par:data}
	When it comes to data, the basic idea is to provide more data through prior knowledge to obtain the desired sample complexity and reduce the estimation risk.
	This can be achieved, e.g., by duplicating training datasets using transformation, by borrowing data from other datasets, or by learning augmentation policies from data as discussed in \cite{CZMVL19}, for more details, refer to Section~\ref{ssec:AtMD}.

\begin{figure}[!h]
		\centering
		\includegraphics[width=0.85\linewidth]{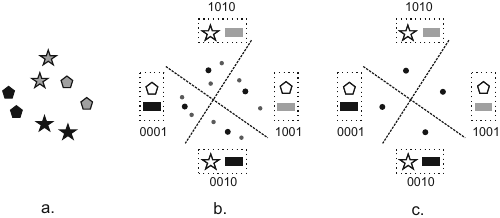}	
		\caption{Prototypical network: a- dataset, b- few-shot learning, c- one-shot learning.}
		\label{fig:proto}
	\end{figure}

	\paragraph{Model-based Methods}\label{par:model}
	In the case of a model-based methods, the primary purpose is to \emph{reduce} sample complexity by restricting the hypothesis space leveraging prior knowledge \cite{WY19} (Fig.~\ref{fig:proto}).
	
	From a probabilistic perspective, much attention has been paid to generative model-based methods \cite{LFP03, STT11, RMDGW16} since prior knowledge plays an important role, although there is a discriminative model-based approach such as \cite{BRSST17}.
	
	\begin{figure}[tbh]
		\centering
		\includegraphics[width=0.9\linewidth]{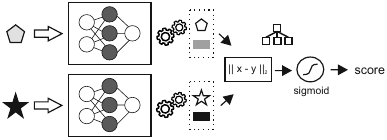}	
		\caption{Siamese Neural Network: The unknown input sample is transformed to output vector and compared with a pre-computed feature vector, the result is a similarity score.}
		\label{fig:siamese}
	\end{figure} 
	
\begin{figure}[tbh]
		\centering
		\includegraphics[width=0.8\linewidth]{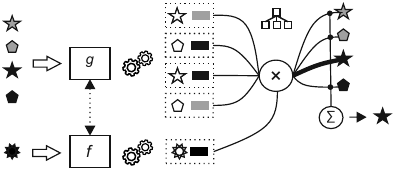}	
		\caption{Matching network (based on \cite[Figure 1]{VBLKW16}): Known and unknown samples use the same (of different) embedding functions (\textit{g} and \textit{f}) for the output vectors, which likelihood is estimated by a given attention mechanism.}
		\label{fig:matching}
	\end{figure}

	Moreover, many NN-based methods have been proposed in recent years by incorporating various architectures of NNs which sometimes leverage external memory, e.g. Siamese NN \cite{KZS15} (Fig.~\ref{fig:siamese}), matching networks \cite{VBLKW16} (Fig.~\ref{fig:matching}), meta-learning \cite{ LBG15, Va18, HAMS20} with memory-augmented NN \cite{SBBWL16a}, multi-task learning \cite{C97, R17} and embedding learning \cite{JSDKLGGD14}.
	
	\paragraph{Algorithm-level Methods}\label{par:algo}
	For designing an appropriate algorithm to solve OSL problems, prior information can help to revise search strategies in hypothesis space in various ways, e.g., providing good initialization, learning search steps, and refining parameters \cite{WY19}. 
	
	Improving OSL methods constitutes a crucial part of human-level AI systems, and thereby, OSL belongs to a very active area of current research \cite{SKP15, DASJSSAZ17, WYKN20}, which has numerous potential applications, among others, object or face recognition \cite{LFP03, CHL05, LFP06, TYRW14}, drug discovery \cite{ARPP17} and detecting home intrusion \cite{CHKXWL17}.
	
	\subsubsection{Surrogate Modeling of Expensive Data}\label{sssec:sur}
	Surrogate modeling is a common approach to handling expensive and often black-box data, a data difficult to analyze and where the mechanisms if their creations is not entirely known.
	
\begin{figure}[H]
		\centering
		\includegraphics[width=0.8\linewidth]{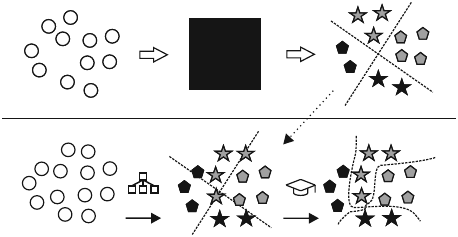}	
		\caption{Surrogate modeling: the data and the predictions from the black-box are used to build surrogate model which is constantly improved in relation of the prediction error.}
		\label{fig:surrogate}
	\end{figure}

	A surrogate presents a simulation or a trained model able to approximate predictions from a black-box model as accurately as possible (Fig.~\ref{fig:surrogate}). The terms often related to surrogate modeling are meta-models, emulators, response surface models, adaptive learning.
	
	To build the prediction function $f({X})$, a large number of samples ${X}$ is required, either too expensive or not available. 
	The observations and their black-box predictions are used to train a simplified and interpretable model as an approximate prediction function $\hat{f}({X})$. 
	The objective is to minimize the prediction error $\epsilon = f({X})-\hat{f}({X})$ with a relatively small number of observations and using a data-driven, bottom-up approach. 
	How good the surrogate model explains the black-box model is measured by the prediction error.
	A surrogate model with acceptable prediction error can interpret the black-box model and estimate the importance of its parameters.
	A new set of observation-prediction pairs can be generated by sampling the input space and enriching the existing dataset.  
	
	Surrogate modeling \cite{FAA08} is commonly used in engineering applications for multi-fidelity optimization of complex dynamic systems and designs; it was also successfully applied in small dataset conditions in biomedical \cite{HKSHHTLA08}, and nonlinear physics applications \cite{TELGF94}.
	
	There is no standard procedure that will select the best surrogate modeling technique for a given problem, and often a poor choice leads to sub-optimal modeling of the dataset. A comprehensive overview of surrogate modeling in engineering design is given by \cite{F08} and \cite{BI18}. For optimization tasks, the choice of new candidate points to be evaluated is crucial and defines the performance and the cost of the surrogate model, at the same time very dependent on the problem. One possible approach is the usage of stochastic radial basis functions with distance weighted selection of minima for new function evaluation with an additional restart capability to avoid the pitfall of getting stuck in a local minimum \cite{RS07}. 
	
	To show a variety of applications in engineering, we refer to aerospace applications \cite{MSGOG17}, usage as flow solver for fluid dynamics \cite{TWPH18}, material science \cite{ZL18, FZD19} or for expensive black-box data \cite{SG10}. In \cite{SLDBHK15} surrogate data was used to find the optimal hyperparameters of ML models built on small datasets (35 and 80 samples).
	
	Recently, physics-informed neural networks (PINN), as data-efficient universal function approximators, have become popular surrogate model framework \cite{RPK19}. PINNs are feed-forward neural networks where structured prior information, usually described by general nonlinear partial differential equations (PDE), is encoded into the learning algorithm.
	The study \cite{DT2022} presents a brief overview of many variants of PINNs and their application used in different fields. More extensive review in \cite{KKLPWY2021} presents some notable applications highlights, such as flow over an espresso cup, 4D flow MRI, edge plasma dynamics, quantum chemistry, material science, geophysics and many others.
	
	\subsection{Approaches to Rare Events}\label{ssec:AtRE}
	In this section, we discuss the possible approaches to the small data challenges regarding rare events as described in Section~\ref{ssec:RE}.
	Rare or low probability events, also known as outliers or anomalies, are different from noise, which arises from random variance. However, rare events could also result from natural variations, changes of system behavior, faulty measurement equipment, or unknown origin. 
	Noise and outliers, which as a result of the expected system behavior, will have a considerable influence on the resulting data model, hence handling of noise is crucial for the identification of real anomalies, and different data pre-processing techniques \cite{JJK11} have to be utilized to achieve robust modeling.
	
	Similar to other ML categories, anomaly detection uses well-known supervised, semi-supervised and unsupervised approaches. 
	
	\subsubsection{Supervised Learning}\label{sssec:SuLe}
\begin{figure}[h]
		\centering
		\includegraphics[width=0.85\linewidth]{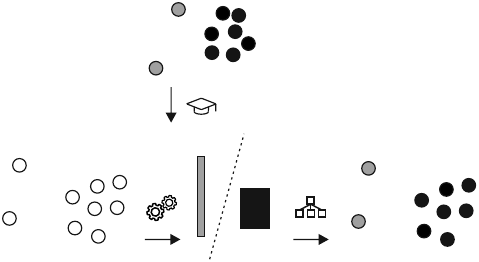}	
		\caption{Supervised anomaly detection.}
		\label{fig:ano1}
	\end{figure}

	In supervised learning, both multi-class and binary classification can be also employed when only the prior information about the normal observations is available. The multi-class classifiers are trained on multiple normal classes; then, an anomalous sample is detected when it is not classified into any of the normal classes using some confidence threshold (Fig.~\ref{fig:ano1}). 
	
	If labeled anomalous samples are available beforehand, then it is possible to train a model treating both classes, anomalous and normal, as part of a standard binary classification task. Usually, this case leverages methods from imbalanced data learning (described in Section~\ref{ssec:ID}), since the anomalous class is mostly the minority class. 
	
	\subsubsection{Semi-supervised Learning}\label{ssec:SSLe}
\begin{figure}[!h]
		\centering
		\includegraphics[width=0.8\linewidth]{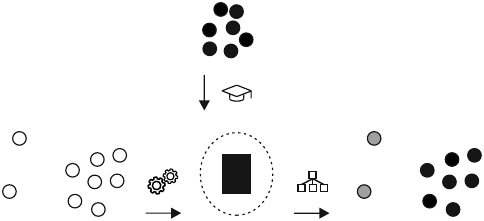}	
		\caption{Semi-supervised anomaly detection.}
		\label{fig:ano2}
	\end{figure}

	In semi-supervised approaches, it is assumed that there are enough labeled observations of the normal data or behavior, and the discriminative boundary can be learned, where any samples outside the boundary are considered anomalous or novel (Fig.~\ref{fig:ano2}).
	
	Common techniques successfully applied in case of small data are: one-class SVMs \cite{SWSWP00, TD04}, isolation forests \cite{LTKZ08}, and recently auto-encoders \cite{B12, CSAT17}. 
	Also, active \cite{GKRB13} and transfer learning could be applied to achieve accurate classification using the least number of labels. 
	
	\subsubsection{Unsupervised Learning}\label{sssec:UnLe}
\begin{figure}[tbh]
		\centering
		\includegraphics[width=0.85\linewidth]{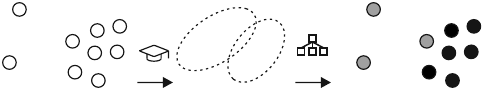}
		\caption{Unsupervised anomaly detection.}
		\label{fig:ano3}
	\end{figure}

	In unsupervised outlier detection, the entire training data set is unlabeled; the only assumption is that the observations considered normal are far more frequent than the anomalous ones (Fig.~\ref{fig:ano3}). 
	The common methods assess the dissimilarities in anomalous observations through the utilization of  statistical, probabilistic, linear, proximity, and deviation models \cite{AC15}.
	
	An extreme-value analysis (EVA) is applied to identify outliers according: the fitness to statistic-probabilistic models, e.g., Gaussian mixture models optimized by Expectation maximization (EM) or reconstruction errors from dimension reduction methods (PCA, autoencoders) \cite{AC15}.
	
	Proximity approaches can be applied as clustering (k-means), statistical (Histogram-based Outlier Score HBOS \cite{GD12}), and the $k$-nearest neighbor methods ($k$-NN\cite{RRS00}, Local Outlier Factor (LOF)\cite{BKNS00} and its variants). A comprehensive overview of unsupervised approaches is given in \cite{GU16}.
	
	There is a wide range of publications about anomaly detection in different application domains \cite{AC15}: cyber-security, financial frauds, medical anomalies, and industrial applications.
	
	As the main application in industrial cases, anomaly detection is state prediction, condition, or health monitoring. 
	Bearing-fault diagnostics in rotary machines \cite{SKK17}, e.g. wind \cite{MS19}, gas-turbines\cite{DJC97, LZ17, YY19}, turbo machines in petroleum industry \cite{MSMG15}, motor bearing fault detection \cite{TMAP16}, then gear-box fault diagnosis in \cite{LYTYPL13, HMKU21} are just few examples among many others.
	
	Major deep learning techniques that have been applied in machine health monitoring are the auto-encoders, RBM, CNN, RNN, and their corresponding variants \cite{ZYCMWG19}. An overview of the ML approaches emphasizing deep learning methods is given in \cite{ZZWH19, PSCH21}.
	
	\section{Conclusion}\label{sec:Concl}
	Machine learning approaches with small data are gaining importance, paralleled by the rising employment of artificial intelligence in industrial and engineering applications.
	
	We categorized small data by five traits: unlabeled data, imbalanced data, missing data, insufficient data, and rare events.  And we introduced methods to cope with different types of small data in industrial applications of machine learning.
	
	We proposed a taxonomy of suitable machine learning approaches and presented current state-of-the-art illustrated by real-world examples for each of the small data challenges. 
	
	\section*{Acknowledgments}
	We thank Ilkin Alkhasli for his comments that greatly improved the manuscript.
	\bibliographystyle{IEEEtran}
	\bibliography{ref}
	\balance
	
\end{document}